\documentclass[11pt,a4paper]{article}
\usepackage[hyperref]{acl2019}
\usepackage{times}
\usepackage{latexsym}
\usepackage{enumitem}
\aclfinalcopy
\usepackage{graphicx}
\usepackage{booktabs}
\usepackage{enumitem}
\newcommand{\club}{\ensuremath\clubsuit}
\newcommand{\spade}{\ensuremath\spadesuit}
\title{Figurative Usage Detection of Symptom Words to Improve Personal Health Mention Detection}
\author{\begin{tabular}{ccc}
\multicolumn{3}{c}{Adith Iyer$^{\spade\club}$,  Aditya Joshi$^{\spade}$, Sarvnaz Karimi$^{\spade}$,  Ross Sparks$^{\spade}$, C\'ecile Paris$^{\spade}$}\\
\end{tabular}\\
\begin{tabular}{ccc}
\multicolumn{3}{c}{$^{\spade}$CSIRO Data61, Sydney, Australia}\\
\multicolumn{3}{c}{$^{\club}$University of Queensland, Brisbane, Australia}\\
\multicolumn{3}{c}{\tt adith.iyer@uq.net.au, \{firstname.lastname\}@csiro.au}
\end{tabular}
}
\begin{document}
\maketitle
\begin{abstract}
Personal health mention detection deals with predicting whether or not a given sentence is a report of a health condition. Past work mentions errors in this prediction when symptom words, \textit{i.e.}, names of symptoms of interest, are used in a figurative sense. Therefore, we combine a state-of-the-art figurative usage detection with CNN-based personal health mention detection. To do so, we present two methods: a pipeline-based approach and a feature augmentation-based approach. The introduction of figurative usage detection results in an average improvement of 2.21\% F-score of personal health mention detection, in the case of the feature augmentation-based approach. This paper demonstrates the promise of using figurative usage detection to improve personal health mention detection.
\end{abstract}

\section{Introduction}
The World Health Organisation places importance on gathering intelligence about epidemics to be able to effectively respond to them~\cite{WHO}. Natural language processing (NLP) techniques have been applied to social media datasets for epidemic intelligence~\cite{charles2015using}. An important classification task in this area is \textit{personal health mention detection}: to detect whether or not a text contains a personal health mention (PHM). A PHM is a report that either the author or someone they know is experiencing a health condition or a symptom~\cite{flu2013}. For example, the sentence `\textit{I have been coughing since morning}' is a PHM, while `\textit{Having a cough for three weeks or more could be a sign of cancer}' is not. The former reports that the author has a cough while, in the latter, the author provides information about coughs in general. Past work in PHM detection uses classification-based approaches with human-engineered features~\cite{flu2013,yin2015scalable} or word embedding-based features~\cite{karisani2018did}. 
However, consider the quote `\textit{When Paris sneezes, Europe catches cold}' attributed to Klemens von Metternich\footnote{\url{https://bit.ly/2VoqTif} ; Accessed on 23rd April, 2019.}. The quote contains names of symptoms (referred to as `\textit{symptom words}' hereafter) `\textit{sneezes}' and `\textit{cold}'. However, it is not a PHM, since the symptom words are used in a figurative sense. Since several epidemic intelligence tools based on social media rely on counts of keyword occurrences~\cite{charles2015using}, figurative sentences may introduce errors. Figurative usage has been quoted as a source of error in past work~\cite{ibm2015,karisani2018did}. In this paper, we deal with the question:
\begin{quote}
\textit{Does personal health mention detection benefit from knowing if symptom words in a text were used in a literal or figurative sense?}
\end{quote}
To address the question, we use a state-of-the-art approach that detects idiomatic usage of words~\cite{idiom}. Given a word and a sentence, the approach identifies if the word is used in a figurative or literal sense in the sentence. We refer to this module as `\textit{figurative usage detection}'. We experiment with alternative ways to combine figurative usage detection with PHM detection, and report results on a manually labeled dataset of tweets. 

\section{Motivation}
\label{sec:motiv}
As the first step, we ascertain if the volume of figurative usage of symptom words warrants such attention. Therefore, we randomly selected 200 tweets (with no duplicates and retweets) posted in November 2018, each containing either `cough' or `breath'. After discarding tweets with garbled text, two annotators manually annotated each tweet with the labels `figurative' or `literal' to answer the question: `\textit{Has the symptom word been mentioned in a figurative or literal manner?}'. Note that, (a) in the tweet `\textit{When it's raining cats and dogs and you're down with a cough!}', the symptom usage is literal, and (b) Hyperbole (for example, `\textit{soon I'll cough my entire lungs up}') is considered to be literal. The two annotators agreed on a label 93.96\% of the time. Cohen's kappa coefficient for inter-rater agreement is 0.8778, indicating a high agreement.  For 52.75\% of these tweets, both annotators assign the label as figurative.  This provides only an estimate of the volume of figurative usage of symptom words. We also expect that the estimate would differ for different symptom words. 

\section{Approach}
We now introduce the approaches for figurative usage and PHM detection. Following that, we present two alternative approaches to interface figurative usage detection with PHM detection: the pipeline approach and the feature augmentation approach.

\subsection{Figurative Usage Detection}
In the absence of a health-related dataset labeled with figurative usage of symptom words, we implement the unsupervised approach to detect idioms introduced in~\citet{idiom}. This forms the figurative usage detection module. The input to the figurative usage detection module is a target keyword and a sentence, and the output is whether or not the keyword is used in a figurative sense. The approach can be summarised in two steps: computation of a literal usage score for target keyword followed by a LDA-based estimator to predict the label. To compute the literal usage score,~\citet{idiom} first generate a set of words that are related to the target keywords (symptom words, in our case). This set is called the `literal usage representation'.  The literal usage score is computed as the average similarity between the words in the sentence and the words in the literal usage representation. Thus, this score is a real value between 0 and 1 (where 1 is literal and 0 is figurative). The score is then concatenated with linguistic features (described later in this section). The second step is a Latent Dirichlet Allocation (LDA)-based estimator. The estimator computes two distributions: the word-figurative/literal distribution which indicates the probability of a word to be either figurative or literal, and a document-figurative/literal distribution which gives a predictive score for a document to be literal or figurative. To obtain the literal usage score, we generate the literal usage representation using word2vec similarity learned from the Sentiment140 tweet dataset~\cite{sentiment140}. We use two sets of linguistic features, as reported in ~\citet{idiom}: the presence of subordinate clauses and part-of-speech tags of neighbouring words, using Stanford CoreNLP~\cite{stanfordcorenlp}. We adapt the abstractness feature in their paper to health-relatedness (\textit{i.e.}, the presence of health-related words). The intuition is that tweets which contain more health-related words are more likely to be using the symptom words in a literal sense instead of figurative. Therefore, the abstractness feature in the original paper is converted to domain relatedness and captured using the presence of health-related words. We consider the symptom word as the target word. It must be noted that we do not have or use figurative labels in the dataset except for the sample used to report the efficacy of figurative usage detection. 
\begin{figure}[tb]
  \centering
            \includegraphics[width=0.52\linewidth]{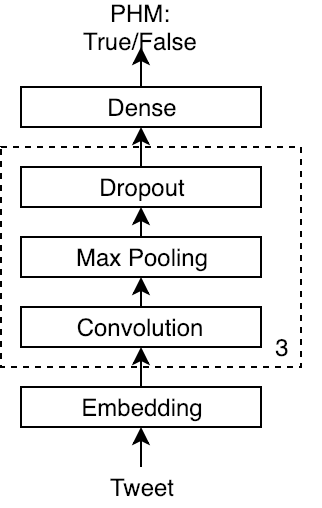}
            \caption{PHM detection.}
            \label{fig:model1aflu}
        \end{figure}
\begin{figure}[t]
    \centering
    \includegraphics[width=0.93\linewidth]{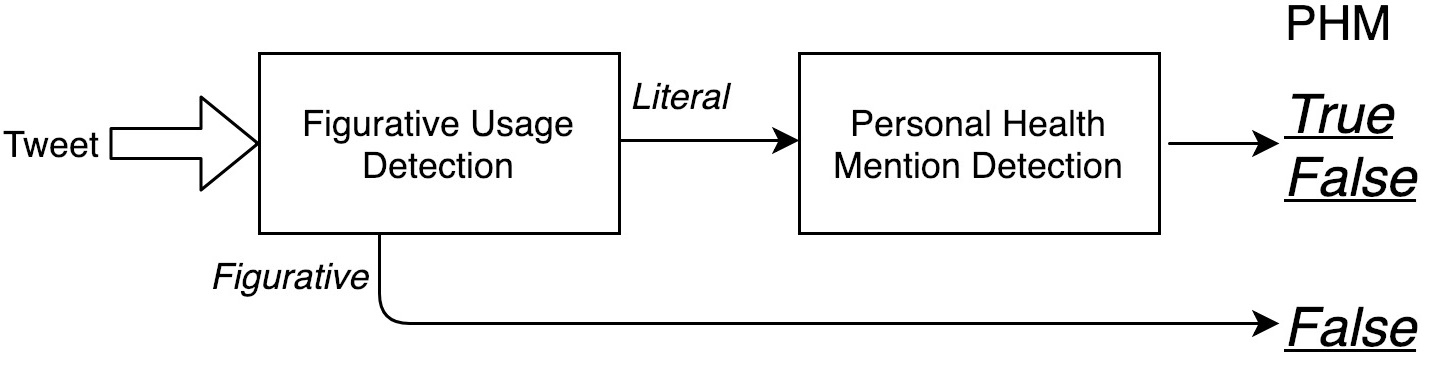}
    \caption{Pipeline approach.}
    \label{fig:discard}
\end{figure}
\subsection{PHM Detection}
We use a CNN-based classifier for PHM detection, as shown in Figure~\ref{fig:model1aflu}. The tweet is converted to its sentence representation using a concatenation of embeddings of the constituent words, padded to a maximum sequence length. The embeddings are initialised based on pre-trained word embeddings. We experiment with three alternatives of pre-trained word embeddings, as elaborated in Section~\ref{sec:expsetup}. These are then passed to three sets of convolutional layers with max pooling and dropout layers. A dense layer is finally used to make the prediction.
\subsection{Interfacing Figurative Usage Detection with PHM Detection}
\begin{figure}[tb]
    \centering
    \includegraphics[width=0.7\linewidth]{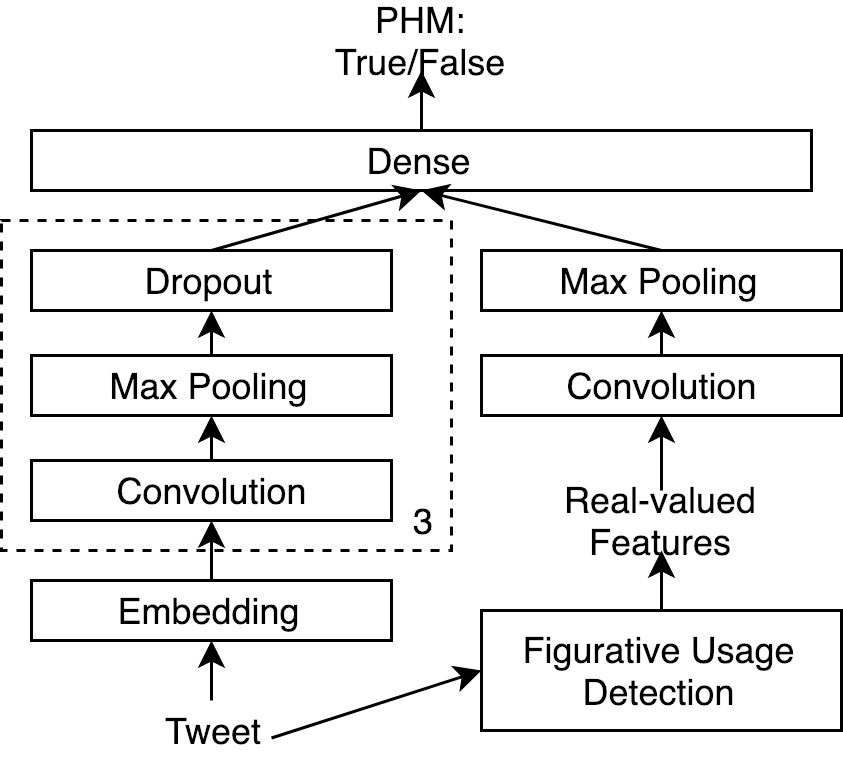}
    \caption{Feature augmentation approach.}
    \label{fig:augment}
\end{figure}

We consider two approaches to interface figurative usage detection with PHM detection:
\begin{enumerate}\setlength \itemsep{0em}
\item \textbf{Pipeline Approach} places the two modules in a pipeline, as illustrated in Figure~\ref{fig:discard}. If the figurative usage detection module predicts a usage as figurative, the PHM detection classifier is bypassed and the tweet is predicted to not be a PHM. If the figurative usage prediction is literal, then the prediction from the PHM detection module is returned. We refer to this approach as `\textit{+Pipeline}'.
\item \textbf{Feature Augmentation Approach} augments PHM detection with figurative usage features. Therefore, the figurative label and the linguistic features from figurative usage detection are concatenated as figurative usage features ad passed through a convolution layer. The two are then concatenated in a dense layer to make the prediction. The approach is illustrated in Figure~\ref{fig:augment}. This approach is based on \citet{dasgupta2018augmenting}, where they augment additional features to word embeddings of words in a document. We refer to this approach as `\textit{+FeatAug}'.
\end{enumerate}
In +Pipeline, the figurative label guides whether or not PHM detection will be called. In +FeatAug, the label becomes one of the features. For both the approaches, the figurative label is determined by producing the literal usage score and then applying an empirically determined threshold. We experimentally determine if using the literal usage score performs better than using the LDA-based estimator (See Section 4.3).

\section{Experiment Setup}
\label{sec:expsetup}
\subsection{Dataset}
We report our results on a dataset introduced and referred to by~\citet{karisani2018did} as the PHM2017 dataset. This dataset consists of 5837 tweets related to a collection of diseases: Alzheimer's (1103, 16.7\% PHM), heart attack (973, 12.4\% PHM), Parkinson's (868, 9.8\% PHM), cancer (988, 20.6\% PHM), depression (924, 38.5\% PHM) and stroke (981, 14.2\% PHM). The imbalance in the class labels of the dataset must be noted. Some tweets in the original paper could not be downloaded due to deletion or privacy settings. 
\subsection{Configuration}
For PHM detection (PHMD) and the two combined approaches (+Pipeline and +FeatAug), the parameters are empirically determined as:
    \begin{enumerate}[leftmargin=*]\itemsep \setlength{0em}
    \item \textbf{PHMD}: Filters=100, Kernels=(3, 4, 5), Pool size=2; Dropout=(0.2, 0.3, 0.5). 
    \item \textbf{Figurative Usage Detection}: The figurative label is predicted using a threshold for the literal usage score. This threshold is set to 0.2. This holds for both +Pipeline and +FeatAug. In the case of +Pipeline, a tweet is predicted as figurative, and, as a result, non-PHM, if the literal usage score is lower than 0.2. In the case of +FeatAug, the figurative label based on the score is added along with other features.
    \item \textbf{+FeatAug}: Filters=100; Kernel size (left)=(3, 4, 5), Pool size=2; Dropout=(0.3, 0.1, 0.3); Kernel size (right)=2. 
\end{enumerate}
\begin{table*}[t]
\centering
\tabcolsep 4pt
\begin{tabular}{llll c lll c lll c lll}
\toprule
& \multicolumn{3}{c}{\bf Random} && \multicolumn{3}{c}{\bf word2vec} && \multicolumn{3}{c}{\bf GloVe} && \multicolumn{3}{c}{\bf Numberbatch} \\
\cmidrule{2-4}\cmidrule{6-8}\cmidrule{10-12}\cmidrule{14-16} 
\textbf{Approach} & \textbf{P} & \textbf{R} & \textbf{F} && \textbf{P} & \textbf{R} & \textbf{F} && \textbf{P} & \textbf{R} & \textbf{F} && \textbf{P} & \textbf{R} & \textbf{F} \\
\midrule
PHMD & 59.40 & 39.84 & 46.31 && 57.85 & 47.24 & 50.99&& 68.71 & 50.70 & 57.05 && 59.07 & 43.59 & 48.63 \\
+Pipeline & 59.99 & 33.65 & 41.78 && 57.84 & 40.80 & 46.62&&67.93 & 43.25 & 51.51&&59.09 & 36.74 & 43.69\\
+FeatAug & 54.51 & 45.01 & \textbf{48.08}&& 57.11 & 51.71 & \textbf{53.13}&& 66.70 & 53.52 & \textbf{58.25} &&54.48 & 48.75 & \textbf{50.45}\\ 
  \bottomrule
   \end{tabular}
\caption{Performance of PHM Detection (PHMD), +Pipeline and +FeatAug for four word embedding initialisations. P:  Precision, R: Recall, and F: F-score.}
\label{tab:phm_cnn}
\end{table*}

\begin{table*}[]
\centering
\tabcolsep 4pt
\begin{tabular}{llll c lll c lll}
\toprule
& \multicolumn{3}{c}{\bf GloVe+MeSH} && \multicolumn{3}{c}{\bf GloVe+WordNet} && \multicolumn{3}{c}{\bf GloVe+Symptom} \\
\cmidrule{2-4}\cmidrule{6-8}\cmidrule{10-12}
\textbf{Approach} & \textbf{P} & \textbf{R} & \textbf{F} && \textbf{P} & \textbf{R} & \textbf{F} && \textbf{P} & \textbf{R} & \textbf{F} \\
\midrule
PHMD & 56.95 & 41.47 & 46.62 && 56.41 & 42.94 & 47.55 && 57.57 & 42.93 & 47.72 \\
+Pipeline &56.01 & 34.98 & 41.75 && 55.86 & 36.63 & 43.12 && 57.10 & 36.34 & 42.90\\
+FeatAug & 53.71 & 46.46 & \textbf{49.01} && 55.88 & 48.47 & \textbf{51.15} && 56.04 & 48.11 & \textbf{50.30} \\
  \bottomrule
   \end{tabular}
\caption{Performance of PHM Detection (PHMD), +Pipeline and +FeatAug initialised with GloVe word embeddings retrofitted with three ontologies: MeSH, WordNet and Symptom. P:  Precision, R: Recall, and F: F-score.}
\label{tab:phm_cnn_retro}

\end{table*}
\begin{table}[tb]
\centering
\begin{tabular}{lllll}\toprule
& \textbf{P} & \textbf{R} & \textbf{F} &  \textbf{$\Delta$F} \\ \midrule
PHMD &  59.48   & 44.18   & 49.26   &               \\ \midrule
+Pipeline & 59.12     &  37.60 & 44.48    & -4.78        \\
+FeatAug & 57.32     & 48.88   & 51.48   & \textbf{+2.21}        \\ \bottomrule
\end{tabular}
\caption{Average performance of PHM Detection (PHMD), +Pipeline and +FeatAug across the seven word embedding initialisations; P: Precision, R: Recall, F: F-score; $\Delta F$: Difference in the F-score in comparison with PHMD.\label{tab:res}}
\end{table}
\begin{table}[h!]
\centering
\begin{tabular}{lll}
\toprule
\textbf{Disease} & \textbf{PHMD} & \textbf{+FeatAug} \\ 
\midrule
Alzheimer's & 65.33& \textbf{68.48}\\
Heart attack & \textbf{46.96}&45.98  \\
Parkinson's & 48.83 & \textbf{51.49} \\
Cancer & 53.69& \textbf{54.58} \\
Depression & 70.48 & \textbf{71.34} \\
Stroke & 57.03 & \textbf{57.65}  \\  
\bottomrule
\end{tabular}
\caption{Impact of figurative usage detection for PHM Detection (PHMD) on individual diseases.\label{tab:phm_indiv}}
\end{table}
All experiments use the Adam optimiser and a batch size of 128, and trained for 35 epochs. CNN experiments use the ReLU activation. We use seven types of initialisations for the word embeddings. The first four are a random initialisation, and three pre-trained embeddings. The pre-trained embeddings are: (a) word2vec~\cite{mikolov2013distributed}; (b) GloVe (trained on Common Crawl)~\cite{pennington2014glove}; and, (c) Numberbatch~\cite{numberbatch}. The next three are embeddings retrofitted with three ontologies. We use three ontologies to retrofit GloVe embeddings using the method by ~\citet{retrofit}. The ontologies are: (a) MeSH,\footnote{\url{https://www.nlm.nih.gov/mesh/meshhome.html}; Accessed on 23rd April, 2019.} (b) Symptom\footnote{\url{https://bioportal.bioontology.org/ontologies/SYMP}; Accessed on 23rd April, 2019.}, and (c) WordNet~\cite{wordnet}. The results are averaged across 10-fold cross-validation.

\subsection{Evaluation of Figurative Usage Detection}
To validate the performance of figurative usage detection, we use the dataset of tweets described in Section~\ref{sec:motiv}. The tweets contain symptom words that have been manually labeled. We obtain an F-score of (a) 76.46\% when only the literal usage score is used, and (b) 69.72\% when the LDA-based estimator is also used. Therefore, we use the literal usage score along with the figurative usage features for our experiments.
\section{Results}
The effectiveness of PHMD, +Pipeline and +FeatAug for the four kinds of word embedding initialisations is shown in Table~\ref{tab:phm_cnn}. In each of these cases, +FeatAug performs better than PHMD, while +Pipeline results in a degradation. We note that, for both +FeatAug and +Pipeline, the recall is impacted in comparison with PHMD.  Similar trends are observed for the retrofitted embeddings, as shown in 
Table~\ref{tab:phm_cnn_retro}. The improvement when figurative usage detection is used is higher in the case of retrofitted embeddings than in the previous case. The highest improvement (47.55\% to 51.15\%) is when GloVe embeddings are retrofitted with WordNet. A minor observation is that the F-scores are lower than GloVe without the retrofitting, highlighting that retrofitting may not always result in an improvement. Table~\ref{tab:res} shows the average performance across the seven types of word embedding initialisations. The +Pipeline approach results in a degradation of 4.78\%. This shows that merely discarding tweets where the symptom word usage was predicted as figurative may not be useful. This could be because the figurative usage detection technique is not free from errors. In contrast though, for +FeatAug, there is an improvement of 2.21\%. This shows that our technique of augmenting with the figurative usage-based features is beneficial. The improvement of 2.21\% may seem small as compared to the prevalence of figurative tweets as described in Section~\ref{sec:motiv}. However, all tweets with figurative usage may not have been mis-classified by PHMD. The improvement shows that a focus on figurative usage detection helps PHMD. Finally, the F-scores for PHMD with +FeatAug with GloVe embeddings for the different illnesses, available as a part of the annotation in the dataset, is compared in Table ~\ref{tab:phm_indiv}. Our observation that heart attack results in the lowest F-score, is similar to the one reported in the original paper. At the same time, we observe that, except for heart attack, all illnesses witness an improvement in the case of +FeatAug. 

\section{Error Analysis}
Typical errors made by our approach are:
\begin{itemize}\setlength \itemsep{0em} 
 \item \textbf{Indirect reference}: Some tweets convey an infection by implication. For example, `\textit{don't worry I got my face mask Charlotte, you will not catch the flu from me!}' does not specifically state that someone has influenza.
 \item \textbf{Health words}: In the case of stroke or heart attack, we obtain false negatives because many tweets do not contain other associated health words. Similarly, in the case of depression, some words like `\textit{addiction}', `\textit{mental}', `\textit{anxiety}' appear which were not a part of the related health words taken into account.
\item \textbf{Sarcasm or humour}: Some mis-classified tweets appear to be sarcastic or joking. For example, `\textit{I'm trying to overcome depression and I need reasons to get out the house lol}'. Here, the person is being humorous (indicated by `\textit{lol}') but the usage of the symptom word `\textit{depression}' is literal.
 \end{itemize}

\section{Related Work}
Several approaches for PHM detection have been reported~
\cite{joshi2019survey}. \citet{flu2013} incorporate linguistic features such as word classes, stylometry and part of speech patterns. \citet{yin2015scalable} use similar stylistic features like hashtags and emojis. \citet{karisani2018did} implement another approach of partitioning and distorting the word embedding space to better detect PHMs, obtaining a best F-score of 69\%. While we use their dataset, they use a statistical classifier while we use a deep learning-based classifier. For figurative usage detection, supervised~\cite{liu2017representations} as well as unsupervised~\cite{sporleder2009unsupervised, idiom, muzny2013automatic, crown} methods have been reported. We pick the work by~\citet{idiom} assuming that it is state-of-the-art.%
\section{Conclusions}
We employed a state-of-the-art method in figurative usage detection to improve the detection of personal health mentions (PHMs) in tweets. The output of this method was combined with classifiers for detecting PHMs in two ways: (1) a simple pipeline-based approach, where the performance of PHM detection degraded; and, (2) a feature augmentation-based approach where the performance of PHM detection improved. Our observations demonstrate the promise of using figurative usage detection for PHM detection, while highlighting that a simple pipeline-based approach may not work. Other ways of combining the two modules, more sophisticated classifiers for both PHM detection and figurative usage detection, are possible directions of future work. Also, a similar application to improve disaster mention detection could be useful (for figurative sentences such as `\textit{my heart is on fire}').

\section*{Acknowledgment}
Adith Iyer was funded by the CSIRO Data61 Vacation Scholarship. The authors thank the anonymous reviewers for their helpful comments. 
\bibliography{bibliography}

\begin{thebibliography}{20}
\expandafter\ifx\csname natexlab\endcsname\relax\def\natexlab#1{#1}\fi

\bibitem[{Charles-Smith et~al.(2015)Charles-Smith, Reynolds, Cameron, Conway,
  Lau, Olsen, Pavlin, Shigematsu, Streichert, Suda et~al.}]{charles2015using}
Lauren Charles-Smith, Tera Reynolds, Mark Cameron, Mike Conway, Eric Lau,
  Jennifer Olsen, Julie Pavlin, Mika Shigematsu, Laura Streichert, Katie Suda,
  et~al. 2015.
\newblock Using social media for actionable disease surveillance and outbreak
  management: A systematic literature review.
\newblock \emph{PloS one}, 10(10):e0139701.

\bibitem[{Dasgupta et~al.(2018)Dasgupta, Naskar, Dey, and
  Saha}]{dasgupta2018augmenting}
Tirthankar Dasgupta, Abir Naskar, Lipika Dey, and Rupsa Saha. 2018.
\newblock \href {https://www.aclweb.org/anthology/W18-3713} {Augmenting textual
  qualitative features in deep convolution recurrent neural network for
  automatic essay scoring}.
\newblock In \emph{Proceedings of the 5th Workshop on Natural Language
  Processing Techniques for Educational Applications}, pages 93--102,
  Melbourne, Australia. Association for Computational Linguistics.

\bibitem[{Faruqui et~al.(2015)Faruqui, Dodge, Jauhar, Dyer, Hovy, and
  Smith}]{retrofit}
Manaal Faruqui, Jesse Dodge, Sujay~Kumar Jauhar, Chris Dyer, Eduard Hovy, and
  Noah~A. Smith. 2015.
\newblock \href {https://doi.org/10.3115/v1/N15-1184} {Retrofitting word
  vectors to semantic lexicons}.
\newblock In \emph{Proceedings of the 2015 Conference of the North {A}merican
  Chapter of the Association for Computational Linguistics: Human Language
  Technologies}, pages 1606--1615, Denver, Colorado. Association for
  Computational Linguistics.

\bibitem[{Go et~al.(2009)Go, Bhayani, and Huang}]{sentiment140}
Alec Go, Richa Bhayani, and Lei Huang. 2009.
\newblock Twitter sentiment classification using distant supervision.
\newblock \emph{CS224N Project Report, Stanford}, 1(12).

\bibitem[{Jimeno~Yepes et~al.(2015)Jimeno~Yepes, MacKinlay, and Han}]{ibm2015}
Antonio Jimeno~Yepes, Andrew MacKinlay, and Bo~Han. 2015.
\newblock \href {https://doi.org/10.18653/v1/W15-3821} {Investigating public
  health surveillance using twitter}.
\newblock In \emph{Proceedings of {B}io{NLP} 15}, pages 164--170, Beijing,
  China. Association for Computational Linguistics.

\bibitem[{Joshi et~al.(2019)Joshi, Karimi, Sparks, Paris, and
  MacIntyre}]{joshi2019survey}
Aditya Joshi, Sarvnaz Karimi, Ross Sparks, Cecile Paris, and C~Raina MacIntyre.
  2019.
\newblock Survey of text-based epidemic intelligence: A computational
  linguistic perspective.
\newblock \emph{arXiv preprint arXiv:1903.05801}.

\bibitem[{Jurgens and Pilehvar(2015)}]{crown}
David Jurgens and Mohammad~Taher Pilehvar. 2015.
\newblock \href {https://doi.org/10.3115/v1/N15-1169} {Reserating the
  awesometastic: An automatic extension of the {W}ord{N}et taxonomy for novel
  terms}.
\newblock In \emph{Proceedings of the 2015 Conference of the North {A}merican
  Chapter of the Association for Computational Linguistics: Human Language
  Technologies}, pages 1459--1465, Denver, Colorado. Association for
  Computational Linguistics.

\bibitem[{Karisani and Agichtein(2018)}]{karisani2018did}
Payam Karisani and Eugene Agichtein. 2018.
\newblock Did you really just have a heart attack?: Towards robust detection of
  personal health mentions in social media.
\newblock In \emph{Proceedings of the World Wide Web Conference}, pages
  137--146, Lyon, France.

\bibitem[{Lamb et~al.(2013)Lamb, Paul, and Dredze}]{flu2013}
Alex Lamb, Michael~J. Paul, and Mark Dredze. 2013.
\newblock \href {https://www.aclweb.org/anthology/N13-1097} {Separating fact
  from fear: Tracking flu infections on twitter}.
\newblock In \emph{Proceedings of the 2013 Conference of the North {A}merican
  Chapter of the Association for Computational Linguistics: Human Language
  Technologies}, pages 789--795, Atlanta, Georgia. Association for
  Computational Linguistics.

\bibitem[{Liu and Hwa(2017)}]{liu2017representations}
Changsheng Liu and Rebecca Hwa. 2017.
\newblock Representations of context in recognizing the figurative and literal
  usages of idioms.
\newblock In \emph{Proceedings of the AAAI Conference on Artificial
  Intelligence}, pages 3230--3236, San Francisco, CA.

\bibitem[{Liu and Hwa(2018)}]{idiom}
Changsheng Liu and Rebecca Hwa. 2018.
\newblock \href {https://www.aclweb.org/anthology/D18-1199} {Heuristically
  informed unsupervised idiom usage recognition}.
\newblock In \emph{Proceedings of the 2018 Conference on Empirical Methods in
  Natural Language Processing}, pages 1723--1731, Brussels, Belgium.
  Association for Computational Linguistics.

\bibitem[{Manning et~al.(2014)Manning, Surdeanu, Bauer, Finkel, Bethard, and
  McClosky}]{stanfordcorenlp}
Christopher~D. Manning, Mihai Surdeanu, John Bauer, Jenny Finkel, Steven~J.
  Bethard, and David McClosky. 2014.
\newblock \href {http://www.aclweb.org/anthology/P/P14/P14-5010} {The
  {Stanford} {CoreNLP} natural language processing toolkit}.
\newblock In \emph{Association for Computational Linguistics (ACL) System
  Demonstrations}, pages 55--60.

\bibitem[{Mikolov et~al.(2013)Mikolov, Sutskever, Chen, Corrado, and
  Dean}]{mikolov2013distributed}
Tomas Mikolov, Ilya Sutskever, Kai Chen, Greg~S Corrado, and Jeff Dean. 2013.
\newblock Distributed representations of words and phrases and their
  compositionality.
\newblock In \emph{Advances in neural information processing systems}, pages
  3111--3119, Lake Tahoe, NV.

\bibitem[{Miller(1995)}]{wordnet}
George~A Miller. 1995.
\newblock Word{N}et: A lexical database for {E}nglish.
\newblock \emph{Communications of the ACM}, 38(11):39--41.

\bibitem[{Muzny and Zettlemoyer(2013)}]{muzny2013automatic}
Grace Muzny and Luke Zettlemoyer. 2013.
\newblock \href {https://www.aclweb.org/anthology/D13-1145} {Automatic idiom
  identification in {W}iktionary}.
\newblock In \emph{Proceedings of the 2013 Conference on Empirical Methods in
  Natural Language Processing}, pages 1417--1421, Seattle, Washington, USA.
  Association for Computational Linguistics.

\bibitem[{Pennington et~al.(2014)Pennington, Socher, and
  Manning}]{pennington2014glove}
Jeffrey Pennington, Richard Socher, and Christopher Manning. 2014.
\newblock \href {https://doi.org/10.3115/v1/D14-1162} {{G}love: Global vectors
  for word representation}.
\newblock In \emph{Proceedings of the 2014 Conference on Empirical Methods in
  Natural Language Processing ({EMNLP})}, pages 1532--1543, Doha, Qatar.
  Association for Computational Linguistics.

\bibitem[{Speer et~al.(2017)Speer, Chin, and Havasi}]{numberbatch}
Robyn Speer, Joshua Chin, and Catherine Havasi. 2017.
\newblock Concept{N}et 5.5: An open multilingual graph of general knowledge.
\newblock In \emph{Proceedings of the Conference on Artificial Intelligence},
  pages 4444--4451, San Francisco, CA.

\bibitem[{Sporleder and Li(2009)}]{sporleder2009unsupervised}
Caroline Sporleder and Linlin Li. 2009.
\newblock \href {https://www.aclweb.org/anthology/E09-1086} {Unsupervised
  recognition of literal and non-literal use of idiomatic expressions}.
\newblock In \emph{Proceedings of the 12th Conference of the {E}uropean Chapter
  of the {ACL} ({EACL} 2009)}, pages 754--762, Athens, Greece. Association for
  Computational Linguistics.

\bibitem[{{World Health Organisation}(2019)}]{WHO}
{World Health Organisation}. 2019.
\newblock Epidemic intelligence - systematic event detection.
\newblock \url{https://www.who.int/csr/alertresponse/epidemicintelligence/en/}.
\newblock [Online; accessed 24-January-2019].

\bibitem[{Yin et~al.(2015)Yin, Fabbri, Rosenbloom, and Malin}]{yin2015scalable}
Zhijun Yin, Daniel Fabbri, S~Trent Rosenbloom, and Bradley Malin. 2015.
\newblock A scalable framework to detect personal health mentions on {T}witter.
\newblock \emph{Journal of Medical Internet Research}, 17(6):e138.

\end{thebibliography}
\bibliographystyle{acl_natbib}
\end{document}